\documentclass[runningheads]{llncs}
\usepackage{wrapfig, booktabs}
\usepackage{graphicx}
\usepackage{subfigure}
\usepackage{amsmath,amssymb}
\usepackage{times}
\usepackage{epsfig}
\usepackage{graphicx}
\usepackage{amsmath}
\usepackage{amssymb}
\usepackage{dirtytalk}
\usepackage{xcolor}
\usepackage{cite}
\usepackage{graphicx}
\usepackage{subfigure}
\usepackage[title]{appendix}
\usepackage{amsmath,amssymb}

\begin{document}
\title{Self-Supervised Discovery of Anatomical Shape Landmarks}
%
%
\author{Riddhish Bhalodia\inst{1,2} \and Ladislav Kavan\inst{2} \and Ross T. Whitaker\inst{1,2} }
%

\institute{Scientific Computing and Imaging Institute, University of Utah
  \and
 School of Computing, University of Utah
}

\maketitle              
\begin{abstract}
Statistical shape analysis is a very useful tool in a wide range of medical and biological applications.  However, it typically relies on the ability to produce a relatively small number of features that can capture the relevant variability in a population. State-of-the-art methods for obtaining such anatomical features rely on either extensive preprocessing or segmentation and/or significant tuning and post-processing. These shortcomings limit the widespread use of shape statistics.  We propose that effective shape representations should provide sufficient information to align/register images. Using this assumption we propose a self-supervised, neural network approach for automatically positioning and detecting landmarks in images that can be used for subsequent analysis.  The network discovers the landmarks corresponding to anatomical shape features that promote good image registration in the context of a particular class of transformations. In addition, we also propose a regularization for the proposed network which allows for a uniform distribution of these discovered landmarks. In this paper, we present a complete framework, which only takes a set of input images and produces landmarks that are immediately usable for statistical shape analysis. We evaluate the performance  on a phantom dataset as well as 2D and 3D images.

\keywords{Self-Supervised Learning, Shape Analysis, Landmark Localization}
\end{abstract}

\section{Introduction}
\label{sec:intro}

Statistical shape modeling (SSM)/morphological analysis \cite{thompson1942growth} is an important resource for medical and biological applications. SSM broadly involves two distinct parts, (i) shape representation which involves describing the anatomy/shape of interest by giving an implicit or explicit representation of the shape, and (ii) using the shape representation to perform the subsequent analysis on shape population. Classical approaches relied on representing the shape via landmark points, often corresponding to distinct anatomical features. There have been many automated approaches for dense correspondence discovery which captures the underlying shape statistics \cite{styner2006spharm, cates2007shape}. An alternate approach to shape representation is to leverage coordinate transformations between images or geometries, typically members of a population or to a common atlas \cite{RTW:Jos2004}. Such a set of transformations implicitly capture the population shape statistics for the objects/anatomies contained in those images.

Automated shape representation via dense correspondences has its drawbacks; most such methods rely on heavily preprocessed data. Such preprocessing steps might include segmentation, smoothing, alignment, and cropping.  These tasks typically require manual parameter settings and quality control thereby making this preprocessing heavy on human resources. In several cases, especially for segmentation a degree of specific anatomical, clinical, or biological expertise is also required, introducing even higher barriers to engaging in shape analysis. Additionally, automated landmark placement or registration rely on computationally expensive optimization methods, and often require additional parameter tuning and quality control. This heavy preprocessing and complex optimization often make statistical shape analysis difficult for nonexperts,  especially when the data under study consists of primarily of images/volumes. 

Systems that produce transformations and/or dense correspondences will typically produce high-dimensional shape descriptors, whereas many users prefer lower-dimensional descriptors to perform subsequent statistical analyses such as clustering, regression, or hypothesis testing. Therefore, there is typically an additional set of processes (e.g. PCA in various forms) that require further expertise (and research) to interpret these complex, high-dimensional outputs and distill them down to usable quantities.  

\begin{figure}
    \centering
    \includegraphics[width=1\linewidth]{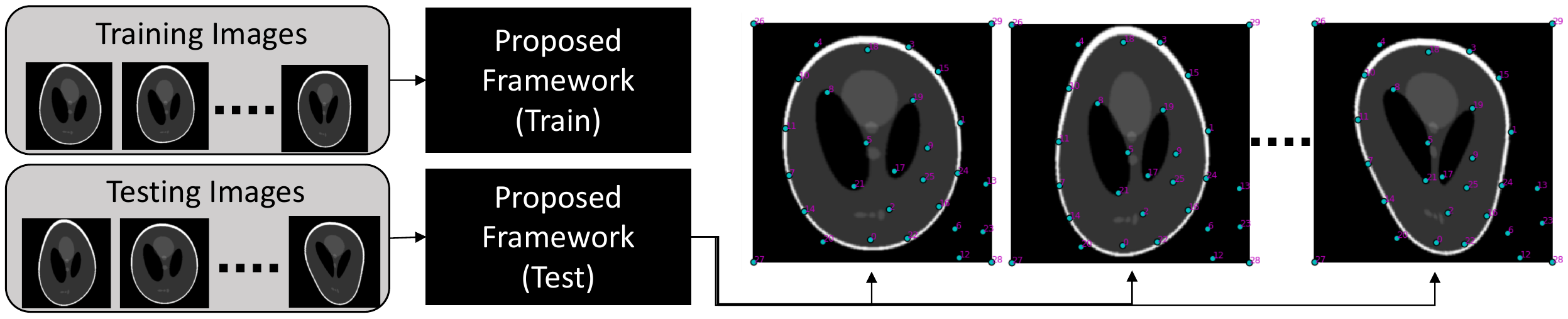}
    \caption{Proposed method's advantage of being an end to end image to landmarks black-box}
    \label{fig:teaser}
\end{figure}


These challenges point a need for an end-to-end system that takes in images and automatically extracts its shape landmarks, for direct statistical shape analysis. In this paper, we propose a system that takes collections of images as input and produces a set of landmarks that leads to accurate registration.  Using image registration as the proxy task, we propose a self-supervised learning method using neural networks that discovers anatomical shape landmarks from a given image set. This method circumvents the intensive preprocessing and optimization required for previous correspondence based methods, while maintaining the simplicity of spatial landmarks of user specified dimension and complexity.
Figure \ref{fig:teaser} showcases the usefulness of the proposed system, i.e. a black-box framework which trains on roughly aligned 2D or 3D images and learns to identify anatomical shape landmarks on new testing images from the same distribution.
\section{Related work}
\label{sec:literature}

Statistical shape models have been extensively used in medical imaging (e.g. orthopedics \cite{harris2013cam}, neuroscience \cite{greig2001brain} and cardiology \cite{cates2013afib}), however, they usually require some form of surface parameterization (e.g. segmentation or meshes) and cannot be applied directly on images. Explicit correspondences between surfaces have been done using geometric parameterizations \cite{RTW:Sty2000, RTW:Dav2002} as well as functional maps \cite{ovsjanikov2012functional}. Particle distribution models (PDMs) \cite{RTW:Gre91} rely on a dense set of particles on surfaces whose positions are optimized to reduce the statistical complexity of the resulting model \cite{cates2007shape, davies2002MDL}. These sets of particles are then projected to a low-dimensional shape representation using PCA to facilitate subsequent analysis \cite{cates2018afib}. Some recent works leverage convolution neural networks (CNNs) to perform regression from images to a shape description of these dense correspondences \cite{Bhalodia2018DeepSSM, bhalodia2018endtoend}. These methods are supervised and require an existing shape model or manual landmarks for their training.

Deformable registration is also widely used as a tool for modeling shape variation/statistics \cite{beg2005computing}, as well as atlas building \cite{RTW:Jos2004}. Recent works rely on neural networks to perform unsupervised registration to atlases \cite{balakrishnan2019tmi} or random pairs \cite{bhalodia2019coop}, and the deformation fields produced by these methods can be used as a shape representation for individuals. Statistical deformation models \cite{rueckert2003nonrigid, joshi1997submanifolds} are applied which learns the probability model for the manifold of deformation fields and reduce the dimension to use it for shape analysis. However, in our experience users across a wide range of applications in medicine and biology prefer the simplicity and interpretability of landmarks.  Many of these same users also balk at the complexity of the dimensionality reduction methods that are needed to make these deformation fields useful for statistical analysis.  

Another relevant work is from computer vision literature, of using an image deformation loss to learn dense features/feature maps \cite{Rocco17geomatch, detone18superpoint}. Other works use convolution neural networks to learn shape features on surfaces to discover correspondences \cite{bronstein2016AnCNN}, or to be used for subsequent correspondence optimizations \cite{agrawal2017deepfeatures}. In this work, we learn landmark points, rather than features/feature maps, in a purely unsupervised setting.
\section{Methods}
\label{sec:methods}

\begin{figure}[!h]
    \centering
    \includegraphics[width=\linewidth]{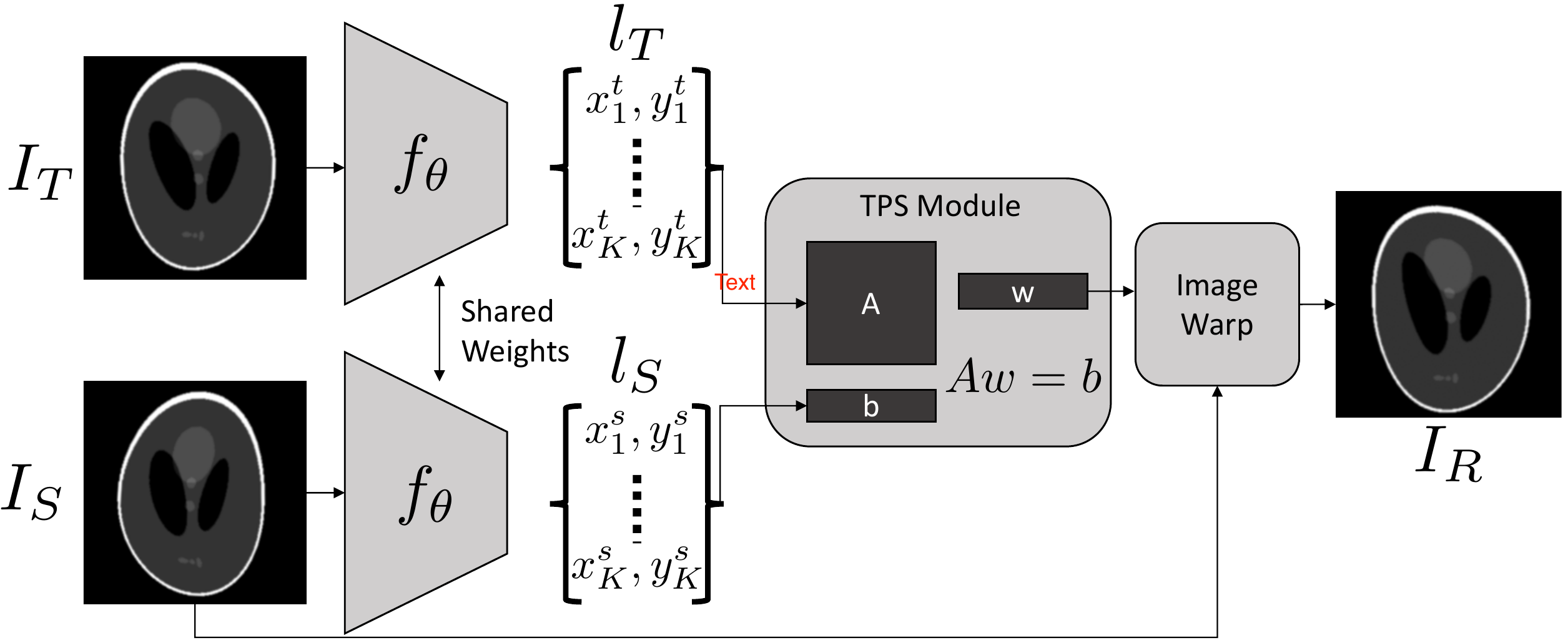}
    \caption{Architecture for discovering anatomical shape landmarks in a self-supervised manner.}
    \label{fig:arch}
\end{figure}

\subsection{Model Description}

The proposed self-supervised approach is that a NN should produce landmarks that best align the objects of interest in pairs of images, using as input many randomly chosen pairs from a set of images. The architecture contains a landmark regressor (NN) which discovers a vector of landmark positions on an image. The architecture includes a spatial transformer stage that takes as input the NN-generated landmarks and compares the resulting aligned images.  For deformation we use a landmark-guided, radial basis function,  \emph{thin plate spline} (TPS) image registration, as in Figure \ref{fig:arch}. For detecting landmarks, we need the system to be invariant pairing, i.e. always produce the same landmarks on an image. Therefore we use Siamese CNN networks \cite{bromley1994signature} to extract the landmarks from source ($I_S$) and the target ($I_T$) images, and we denote these points $l_S$ and $l_T$ respectively.  A single instance/sibling of the fully trained Siamese pair is the {\em detector}, which can be applied to training data or unseen data to produce landmarks.   The output of a network (from the Siamese pair) is a $K \times d$ matrix, where $K$ is the number of landmark points and $d$ is the dimensionality of the image, typically either 2D or 3D. These features are treated as correspondences that go into the TPS solve module. The TPS registration entails the following steps: (i) create the kernel matrix $A$ from $l_T$ and the output position vector $b$ using $l_S$, (ii) solve the linear system of equations ($Aw = b$) to get the TPS transformation parameters $w$, and (iii) using $w$ the source image, warp the entire source grid into the registered image ($I_R$). 

The loss for the entire network comprises of two terms:
\begin{align}
    \mathcal{L} = \mathcal{L}_{match}(I_T, I_R) + \lambda\mathcal{L}_{reg}(A)
\end{align}
The first term is the image matching term, any image registration loss which allows for backpropagation can be employed, in this work we work with $\mathbb{L}$-2. However, more complex losses such as normalized cross-correlation \cite{balakrishnan2019tmi} can also be used. The second term is the regularization term applied on the TPS transformation matrix kernel $A$, so that it does not become ill-conditioned, which can happen from the random initialization of the network or during optimization. To regularize $A$ we use the condition number of the matrix using Frobenius Norm ($||A||_F = \sqrt{\sum\limits_{i=1}^M\sum\limits_{j=1}^N a^2_{ij}}$), defined as follows:
\begin{align}
    \mathcal{L}_{reg}(A) = \kappa_F(A) = ||A||_{F}||A^{-1}||_{F} 
\end{align}
 A useful side effect of this regularization is as follows.   The TPS kernel matrix becomes singular/ill-conditioned when two landmark positions are very close together, and therefore this regularization term, while preventing the matrix from being singular also encourages the resulting shape landmarks to be distributed in the image domain. This regularization loss is weighted by a scalar hyperparameter $\lambda$, and our experience is that the system produces useful and effective landmarks for a wide range of $\lambda$'s.  

\subsection{Training Methodology}
\label{subsec:training}
The network is trained on a set of 2D/3D images in a pair-wise fashion.  For smaller data sets, it is feasible to present the network with all pairs of images, as we have in the examples here.  For larger image datasets, the $n^2$ training size may be too large and/or redundant, and random pairs could be used.   For robust training, we train the Siamese network on images with a small added i.i.d Gaussian noise on the input, while using the original images for the loss computation on the transformed image. Finally, we whiten the image data (with zero mean and unit variance) before feeding it into the networks. 

\begin{figure}[!h]
    \centering
    \includegraphics[width=\linewidth]{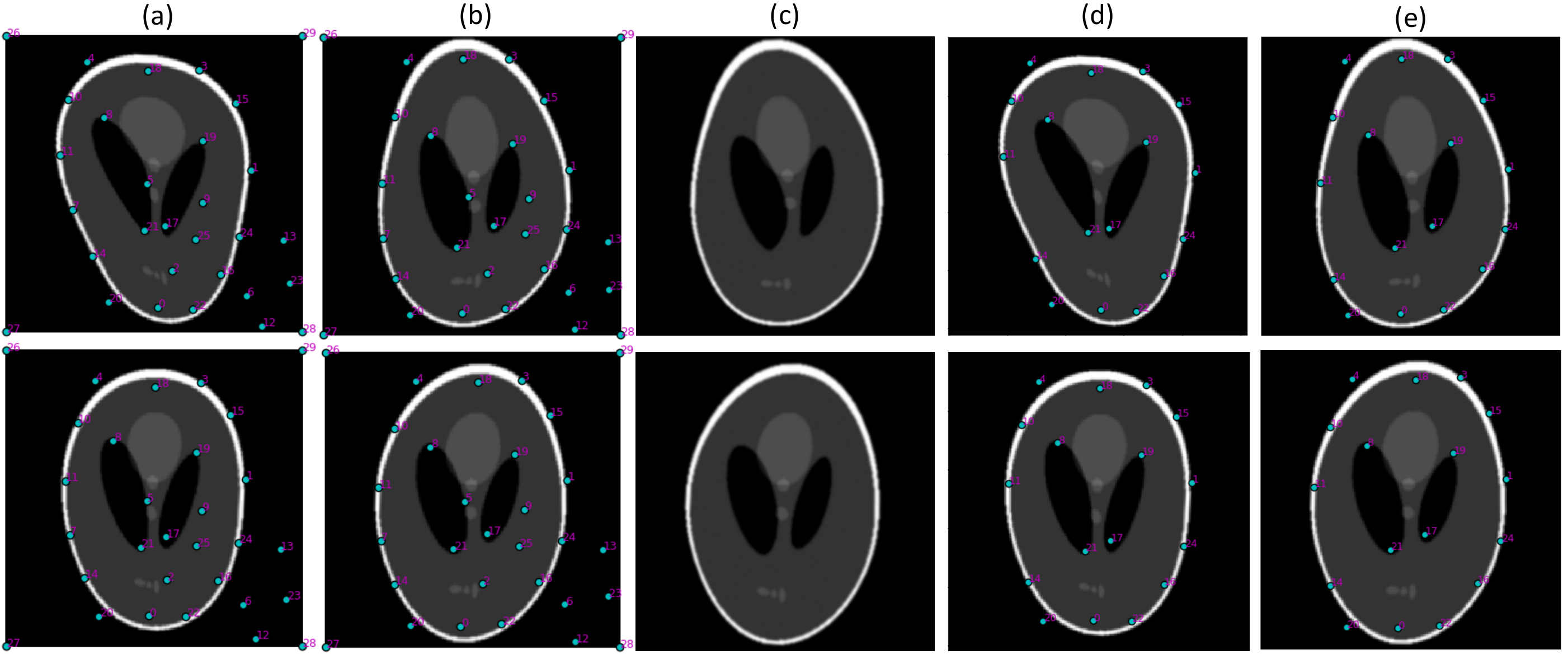}
    \caption{Results on S-L phantom dataset, (a) and (b) an example of source-target image pair with learned landmarks overlayed, (c) the registered image (source image warped to target) and, (d) and (e) the same source and target pair with landmarks after the redundant points removal.}
    \label{fig:phantomFig}
\end{figure}

\textbf{Landmarks removal :} As it can be seen in Figure \ref{fig:teaser}, when we discover landmarks there may be some which are \emph{redundant}, i.e. they do not align with any significant anatomical features important for image registration. For this we propose a post-training method to weed out the redundant landmarks. After training the network, we reregister pairs of images by removing one landmark at a time, and we compute the difference in the registration loss computed using all the landmarks and by leaving one of them out. Averaging over all image pairs gives us an indicator of the importance of each landmark, and we can threshold and remove the ones with low importance. We apply this technique to the phantom data for testing.  Figure \ref{fig:phantomFig} shows images/landmarks with and without the redundancy removal, demonstrating that landmarks in non-informative regions are removed.   There are many alternative methods for culling landmarks, such as L1 regularized weights that the network might learn.  We consider the study of alternative methods for landmark culling as beyond the scope of this paper.  

\section{Results}
\label{sec:results}
In this section we describe the performance of the proposed method on three datasets, (i) a Shepp-Logan (S-L) phantom (sythnthetic) dataset,  (ii) a diatoms dataset with 2D images of diatoms of four morphological classes, (iii) a dataset of 3D cranial CT scans of infants for detection of craniosynostosis (a morphological disorder). We split each dataset into training, validation and testing images with 80\%, 10\%, 10\% division, and the image pairs are taken from their respective set. Detailed architecture for each dataset is presented in the supplementary material.

\subsection{Phantom Dataset}
\label{subsec:phantom}
We create a synthetic dataset using S-L phantom and creating randomly perturbed TPS warps using 6 control points, 2 on each ellipse endpoints, the boundary ellipse, and 2 black ellipses inside the phantom.  We train  our network
 on this dataset with 30 landmarks, with 4 constant landmarks placed on the corners of each image. We train the proposed network on this dataset for 20 epochs with a regularization coefficient of 0.0001, this gives an average registration loss to reach ~0.01\%. Figure \ref{fig:phantomFig} showcases the results on this dataset, \ref{fig:phantomFig}(a) and \ref{fig:phantomFig}(b) are the source and target images overlayed with predicted features respectively, \ref{fig:phantomFig}(c) represents the registered output. We also remove the redundant points as described in section \ref{subsec:training}, the retained points are shown in Figure \ref{fig:phantomFig}(d, e). Encouragingly, the redundancy metric shows high importance for points that are near the location of the control points from which the data was generated.

\subsection{Diatoms Dataset}
\label{subsec:diatoms}
\begin{figure}
    \centering
    \includegraphics[width=\linewidth]{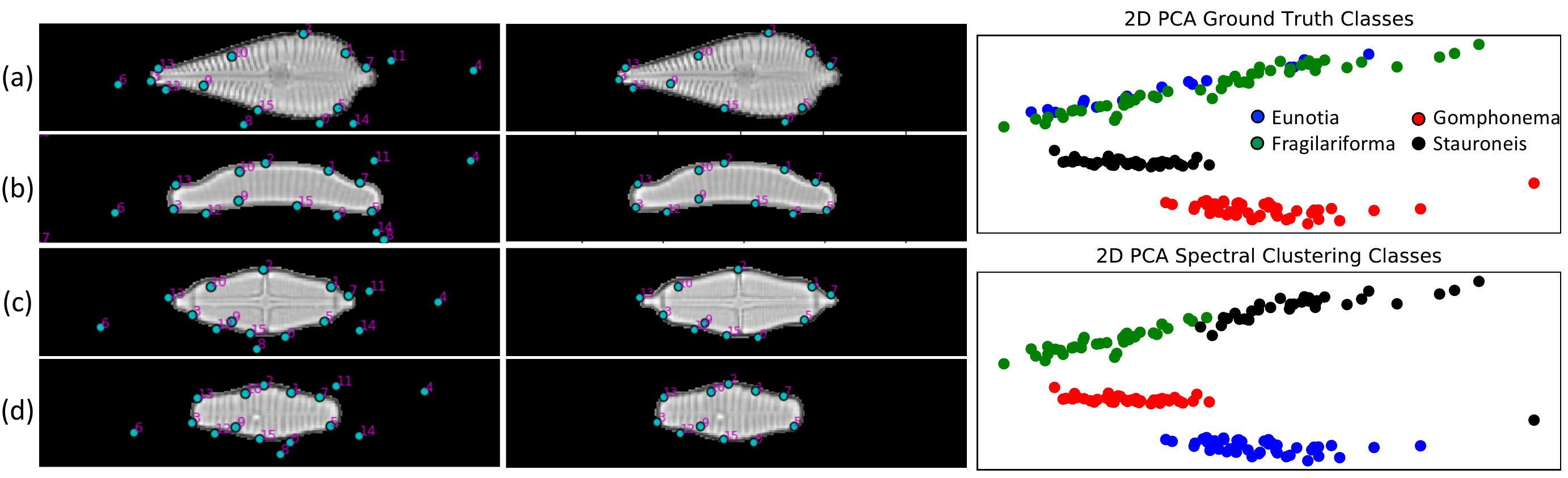}
    \caption{ The left image rows( (a) Gomphonema, (b) Eunotia, (c) Stauroneis, (d) Fragilariforma ) showcase 4 different images from test set corresponding to four different classes of diatoms and their predicted landmarks shown in cyan. The middle column shows the same images with redundant landmarks removed. The left column shows spectral clustering (in 2D PCA embedding space) evaluated on a test set and compared to ground truth labels.}
    \label{fig:diatoms}
\end{figure}

Diatoms are single-cell algae which are classified into different categories based on their shape and texture, which have biological applications \cite{hicks2006diatoms}. The dataset consists of 2D diatom images with four different classes \emph{Eunotia} (68 samples), \emph{Fragilariforma} (100 samples), \emph{Gomphonema} (100 samples) and \emph{Stauroneis} (72 samples). We train the proposed network on a joint dataset of images from all four classes (all together, unlabeled/unsupervised), allowing the network to establish correspondences between different classes via discovered landmark positions.  The goal is to see how well the discovered landmark positions describe the shape variation and distinguish the classes in the unsupervised setting.  We whiten(zero mean, unit variance) the images before passing it through the network, use the regularization parameter set to 1e-5, and train the network to discover 26 landmarks. We train this dataset for 20 epochs which achieves the testing registration accuracy of ~0.1\%. Like with phantom data we compute the redundancy metric and remove landmarks below a threshold. Figure \ref{fig:diatoms} (middle column) shows the images with only the retained landmarks and we see that the landmarks which did not lie on shape surfaces (left column) were removed for lack of impact on the losss function (point numbers 4, 6, 8, 14). 
We want to evaluate how well these landmarks used as shape descriptors perform on clustering these diatoms. For this, we use Spectral clustering \cite{Luxburg07spectral} on the learned landmarks.  We cluster landmarks coming from the training images and then predict the clusters for the validation as well as testing images (using the landmarks). The results are shown in the right column, which shows 2D PCA embedding of landmarks (shape features) labeled with ground truth clusters labels and the predicted cluster label. The mismatch between \emph{Eunotia} and  \emph{Fragilariforma} classes is due to the fact they both have a very similar shapes/sizes are typically also distinguished by intensity patterns, which are not part of this research. 

\subsection{Metopic Craniosynostosis}
\begin{figure}[!h]
    \centering
    \includegraphics[width=\linewidth]{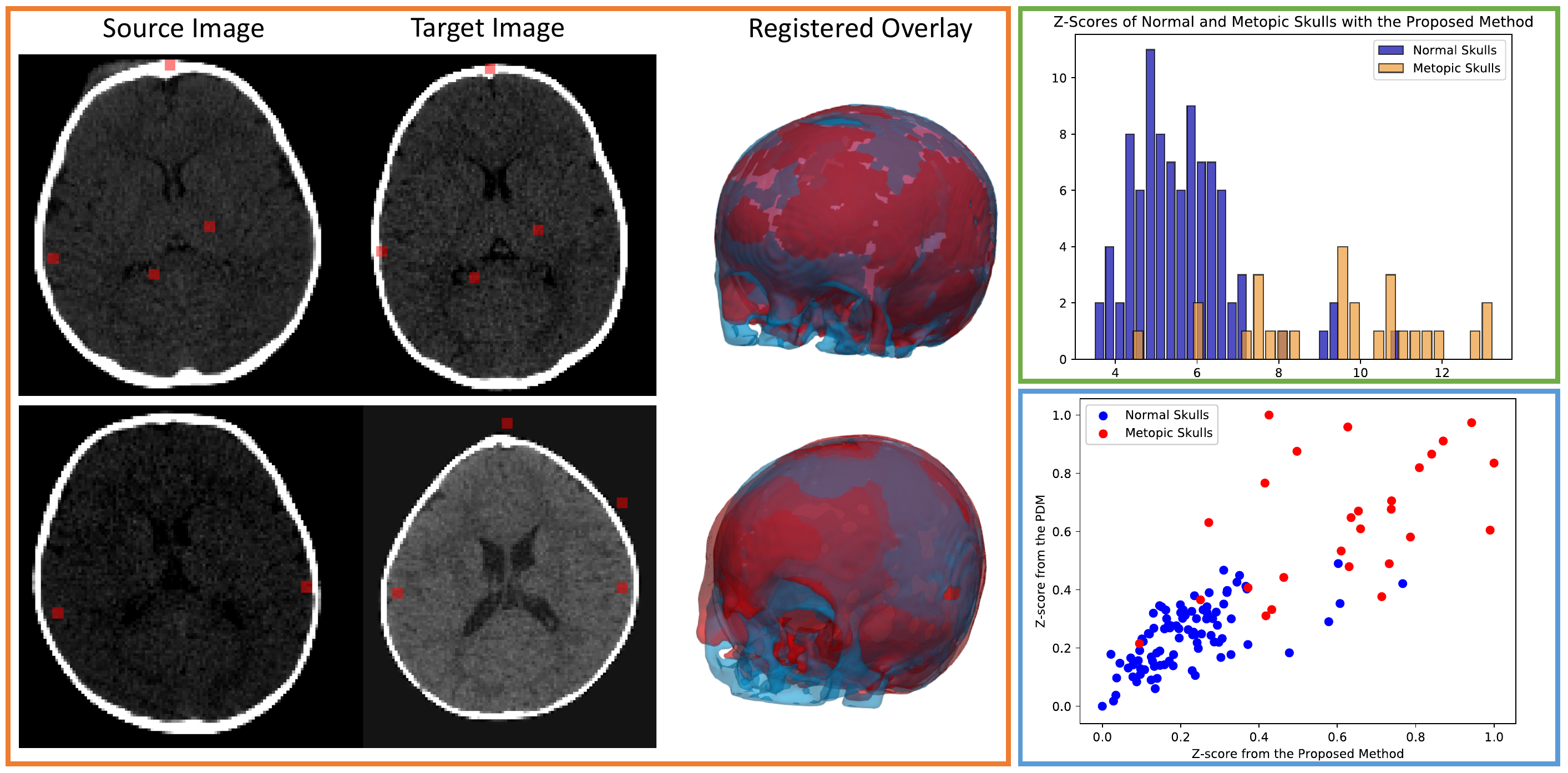}
    \caption{(Orange Box) represents two instances of source and target 3D volumes, the first two images also show a specific contrast enhanced axial slice with discovered landmarks, and the third figure shows overlay of target segmentation and the registration segmentation. (Blue Box) Correlation between Z-scores from state-of-the-art PDM and Z-scores of landmark shape descriptors from the proposed method. (Green Box) Showcases the Z-scores from the landmark shape descriptor which is able to distinguish normal versus metopic skulls.}
    \label{fig:cranio}
\end{figure}

Metopic craniosynostosis is a morphological disorder of cranium in infants, caused by premature fusion of the metopic suture. The severity of metopic craniosynostosis is hypothesized to be dependent on the deviation of a skull's morphology from the shape statistics of the normal skull. There have been studies on diagnosing this condition \cite{wood2016name} as well as characterizing the severity of the pathology \cite{bhalodia2019severity} by various shape representation and modeling techniques. For this example we use 120 CT scans of the cranium with 93 normal phenotypical skulls with no deformity and 27 samples of metopic craniosynostosis (pathological scans for such conditions are generally less abundant). Each CT scan is a $101 \times 118 \times 142$ volume with an isotropic voxel resolution of 2mm. We whiten (zero mean, unit variance) the data for training the network and train for 10 epochs with regularization weight of 0.00001 and 80 3D landmarks. The network saw both classes of scans but with no labels/supervision.  After the redundancy removal we are only left with 49 landmarks. The registration performance of the system is shown in Figure \ref{fig:cranio} (orange box) at the mid-axial slice of two source-target pairs. CT scans do not present with apparent information inside the skulls, however, with a very specific contrast setting (used in the figures) we can see grainy outlies of brain anatomy inside the skulls. An interesting thing to note is certain landmarks adhere to these low-contrast features, and are consistent across shapes. In row two of the figure, we show a normal skull as the source image and the target is a metopic skull, and trigonocephaly (triangular frontal head shape) which is a key characteristic in metopic craniosynostosis is captured by the three (in plane) landmarks, one on the frontal rim and two near the ears. Capturing this morphological symptom is essential in distinguishing the normal skulls from the metopic.
 Using these 49 landmarks as the shape descriptors we evaluate how well it can capture the deviation between the metopic skull from the normal population. 

\textbf{Severity Comparison :} First, we want to study if the discovered landmarks are descriptive enough to accurately identify the pathology. We use all 120 images (irrespective of the training/testing set), we compute Mahalanobis distance (Z-score) of all the images from the base distribution formed using the normal skulls in the training dataset. Figure \ref{fig:cranio}(green box) is the histogram of the Z-scores, it showcases that the Z-score of metopic skulls is larger on average than the normal skulls which support our hypothesis that the discovered low-dimensional landmark-based shape descriptor is enough to capture abnormal morphology. We also want to compare against a shape-of-the-art, correspondence-based shape model. We use \emph{ShapeWorks} \cite{cates2007shape} as our PDM which involved segmenting the skulls, preprocessing (smoothing, anti-aliasing, etc.) and nonlinear optimizations, to place 2048 particles on each skull. \emph{ShapeWorks} has been used in other metopic craniosynostosis severity analyses such as in \cite{bhalodia2019severity}.
We perform PCA to reduce the dense correspondences to 15 dimensions (95\% of variability) and compute Z-scores (the base data to be the same scans as with the proposed method). We normalize Z-scores from both methods (PDM and the proposed one) and plot their scatter plot in Figure \ref{fig:cranio} (blue box). We see that the correlation between the two methods is significant (correlation coefficient of 0.81), and hence, can predict/quantify the severity of metopic craniosynostosis equally well. This demonstrates that the landmarks discovered by our method are a good substitute for the state-of-the-art PDM model with very low amounts of  pre- and postprocessing overhead. Also, it is much faster to get shape descriptors for new images as compared with conventional PDM methods.

\section{Conclusions}
\label{sec:conclusions}
In this paper, we present an end-to-end methodology for processing sets of roughly aligned 2D/3D images to discover landmarks on the images, which then can be used for subsequent shape analysis. Under the assumption that good landmarks can produce a better image to image registration, we propose a self-supervised neural network architecture that optimizes the TPS registration loss between pairs of images with an intermediate landmark discovery that feeds into the TPS registration module. Additionally, we also propose a regularizer that ensures TPS solvability and encourage more uniform landmark distribution. We also apply a redundancy removal via the loss difference with and without a particular particle, which manages to cull uninformative landmarks. However, for future work there is a possibility of assigning importance to individual landmarks during training which can be used for removing redundant points. This paper presents a combination of technologies that result in a turn-key system, that will allow shape analysis to be used in a wide range of applications with a much less expertise than previous methods.  We showcase that our methodology places landmarks qualitatively corresponding to distinct anatomical features and that these landmarks can work as shape descriptors for downstream tasks such as clustering for diatoms or severity prediction for metopic craniosynostosis.
\appendix
\bibliographystyle{splncs04}
\bibliography{cranio}
\section*{Appendix}
\begin{figure}
    \centering
    \includegraphics[width=\linewidth]{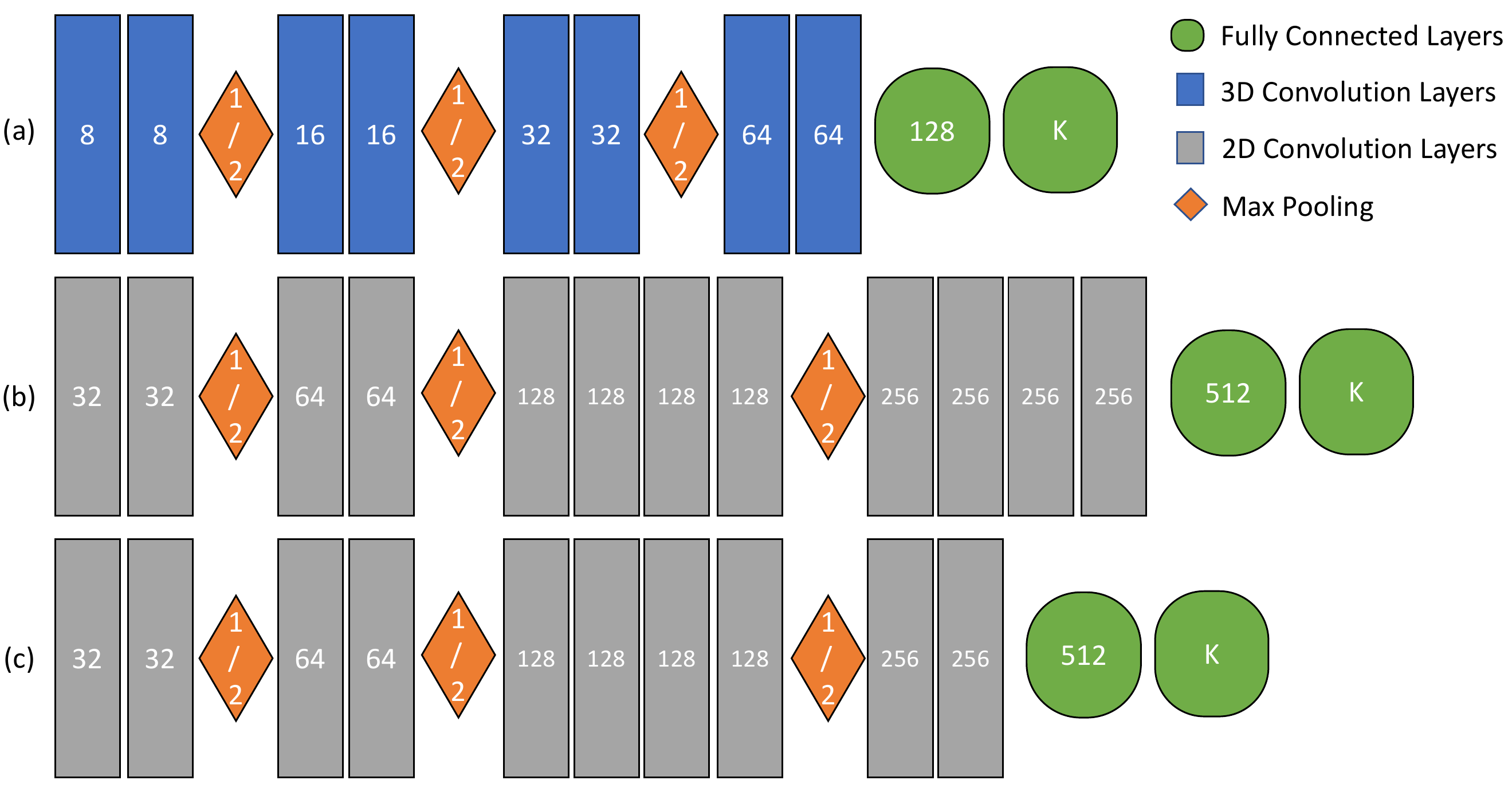}
    \caption{(a) Network architecture for metopic craniosynostosis 3D volumes to landmark positions, (b) network architecture for phantom images, (c) network architecture for diatom images. The numbers in each of the convolution and fully connected block specifies the output number of channels.}
    \label{fig:det}
\end{figure}

The detailed architecture for each of the landmark regressor is given in Figure \ref{fig:det}. In each of the convolution layers we use the kernel size of $3\times3$ for 2D and $3\times3\times3$ for 3D. The non-linear activation used are ReLU's and hyperbolic tangent is used for the output layer. 

\end{document}